\pdfoutput=1

\documentclass[11pt]{article}

\usepackage{EMNLP2022}

\usepackage{amsmath}
\usepackage{amsthm}
\usepackage{times}
\usepackage{bm}
\usepackage{txfonts}
\usepackage{xspace}
\usepackage{graphicx}
\usepackage{amsfonts}
\usepackage{bm}
\usepackage{booktabs}
\usepackage{algorithm}
\usepackage{algorithmic}
\usepackage{latexsym}
\usepackage{array}
\usepackage{multirow}
\usepackage{subcaption}
\usepackage{enumitem}

\usepackage{fancyhdr}

\usepackage[capitalize]{cleveref}
\usepackage[misc]{ifsym}

\Crefname{equation}{Eq.}{Eqs.}
\Crefname{figure}{Fig.}{Figs.}

\DeclareMathOperator*{\argmax}{arg\,max}

\usepackage[T1]{fontenc}

\newcommand{\correct}[1]{\textcolor[RGB]{0,100,0}{#1}}
\newcommand{\ourframework}{{\sc TASD}\xspace}
\newcommand{\ourmodel}{{\sc TASATG}\xspace}

\usepackage[utf8]{inputenc}

\usepackage{microtype}

\title{
Towards Table-to-Text Generation with Pretrained Language Model: A Table Structure Understanding and Text Deliberating Approach
}

\author{
  Miao~Chen{\textsuperscript{$\S\, $}}{\thanks{~~This work was done when the first author was an intern at Baidu Research under the supervision of the second author.}}~,
  Xinjiang~Lu{\textsuperscript{$\P\,\,{\textrm{\Letter}}$}},
  Tong~Xu{\textsuperscript{$\S$}},
  Yanyan~Li{\textsuperscript{$\,\P$}},
  Jingbo~Zhou{\textsuperscript{$\P$}}, 
  Dejing~Dou{\textsuperscript{$\P$}},
  Hui~Xiong{\textsuperscript{$\dagger\,\ddagger$}}
  \\
  \textsuperscript{$\P$}BIL, Baidu Research,~~
  \textsuperscript{$\S$}University of Science and Technology of China 
  \\
  \textsuperscript{$\dagger$}Hong Kong University of Science and Technology (Guangzhou) \\
  \textsuperscript{$\ddagger$}Guangzhou HKUST Fok Ying Tung Research Institute \\
  \texttt{
  cmer@mail.ustc.edu.cn,
  $\{$%
  luxinjiang,liyanyanliyanyan,zhoujingbo%
  $\}$@baidu.com,
  } \\
  \texttt{
  tongxu@ustc.edu.cn,
  doudejing@baidu.com,
  xionghui@ust.hk
  } \\
  }

\begin{document}

\maketitle
\thispagestyle{fancy}

\begin{abstract}


Although remarkable progress on the neural table-to-text methods has been made, the generalization issues  hinder the applicability of these models due to the limited source tables. 
Large-scale pretrained language models sound like a promising solution to tackle such issues. 
However, how to effectively bridge the gap between the structured table and the  text input by fully leveraging table information to fuel the  pretrained model is still not well explored.~%
Besides, another challenge of integrating the deliberation mechanism into the text-to-text pretrained model for solving the table-to-text task remains seldom studied. 
In this paper, to implement the table-to-text generation with pretrained language model, we propose a \underline{ta}ble \underline{s}tructure understanding and text \underline{d}eliberating approach, namely \ourframework. 
To be specific, we devise a three-layered multi-head attention network to realize the table-structure-aware text generation model with the help of the pretrained language model. 
Furthermore, a multi-pass decoder framework is adopted to enhance the capability of polishing generated text for table descriptions.
The empirical studies, as well as human evaluation, on two public datasets, validate that our approach can generate faithful and fluent descriptive texts for different types of tables.

\end{abstract}

\section{Introduction}

The task of learning to generate natural language descriptions from non-linguistic input, which is referred to as data-to-text, is important for many applications, such as weather forecast generation~\citep{mei2016talk}, sports news writing~\citep{wiseman2017challenges}, biography writing~\citep{lebret2016neural}, market comments writing~\citep{murakami2017learning} and automatic question-answering~\citep{li2021tsqa}. 
The input data can be in various forms for data-to-text though, here we focus on the text generation task that takes the table as input.

Inspired by neural machine translation models, previous studies on table-to-text tasks mainly adopt traditional seq2seq methods to generate table  descriptions~\citep{lebret2016neural,wiseman2017challenges,liu2018table,gong2019enhanced,wang2020towards,li2021improving}. 
Despite generating text with high fluency, 
lacking numerous source tables leads to lower generalizability of the table-to-text model.
Recent progress in the pretrained language model \citep{kenton2019bert,radford2019language} shows remarkable performance in solving natural language processing tasks. 
The model pretrained on large-scale data possesses rich knowledge, which inspires us with the potential for solving generalization issues of the text generation task.


To exploit the expressive power of the pretrained model for the table-to-text task, it is necessary to serialize the input table effectively. 
Several works have put efforts to bridge this gap, such as serializing the table into a token sequence~\citep{zhang2020table,suadaa2021towards,xing2021structure}, or introducing an extra task to control the table representation~\citep{gong2020tablegpt}. 
However, none of these leveraged the table structure information effectively. 
Furthermore, the text-to-text pretrained model decodes and generates a sequence in a one-pass forward process, which means it cannot perceive the future words in advance on the target side. 
Recently, the deliberation mechanism~\citep{niehues2016pre,geng2018adaptive} implemented by the multi-pass decoder is proposed to tackle this problem.
However, how to adapt this approach for text-to-text pretraining, which can be further applied to the table-to-text task, is another challenge.

To this end, we propose a \underline{ta}ble \underline{s}tructure understanding and text \underline{d}eliberating approach, namely \ourframework, to solve the table-to-text task with the pretrained language model enhanced by the deliberation mechanism. 
Specifically, we first serialize the table input with customized templates which do not acquire the target cells to be labeled. 
Then, we employ the multi-head attention in a hierarchical way to learn the table representation that is aware of table structure and apply it to guide the fine-tuning of the text-to-text pretrained model. 
Afterward, we adopt the multi-pass decoder to realize text deliberation. 
More specifically, we treat the above table-structure-aware fine-tuned model as the first-pass decoder and adopt another pretrained model as the second-pass decoder to further polish the descriptive text. 
In the second-pass decoding phase, the table representation can be conveniently leveraged as the ``original text'' in the text deliberation mechanism. 
The main contributions of this work can be summarized as follows: 

\vspace{-1.5ex}
\begin{itemize}

\item 
We propose a novel table-to-text generation approach (i.e., \ourframework) to assimilating the complete table information with the help of table structure distillation, the pretrained language model, and the text deliberation. 

\vspace{-1ex}
\item 
We devise a table-structure-aware text generation model (\ourmodel) via the hierarchical multi-head attention network, which can realize the content selection automatically.
And we develop an effective text deliberation method dedicated to the table-to-text task.  

\vspace{-1ex}
\item 
Extensive experiments conducted on two different datasets demonstrate that \ourframework ~outperforms comparable baselines in terms of various metrics. 

\end{itemize}

\vspace{-2ex}
\section{Related Work}

\vspace{-1ex}
\subsection{Table-to-Text Generation}
\vspace{-.5ex}

Encouraged by the success of seq2seq methods in machine translation and text summarization,  researchers proposed to formulate the input table as a sequence of records \cite{lebret2016neural,wiseman2017challenges}, and further improve the performance of table-to-text methods based on seq2seq by modeling table representation \citep{liu2018table,gong2019table}.~%
Introducing auxiliary tasks to enrich the table representation~\citep{tian2019sticking,li2021improving} is another promising paradigm to address the table-to-text problem.~%
Moreover, there have been studies focusing on how to disaggregate the table-to-text pipeline effectively to generate more faithful and fluent text, e.g. leveraging content selection and planning~\citep{puduppully2019data,trisedya2020sentence,bai2021infobox}, combining autoregressive and non-autoregressive methods~\citep{wang2021sketch}. 
In addition, recent Transformers were also applied to solve the table-to-text task~\citep{gong2019enhanced,wang2020towards,obeid2020chart}. 
However, current table-to-text methods may fail to tackle the overfitting problem aroused by the lack of diversity in small datasets.

Fine-tuning the model pretrained in a large corpus and adapting to a specific task is an effective approach to tackling the generation issues disturbed by small data and large parameters~\citep{radford2019language}. 
\citep{kale2020text} explored the feasibility of applying the text-to-text pretrained model to the table-to-text task, 
\citep{gong2020tablegpt} applied multi-task learning to solve the table-to-text task with pretrained language model,  and \citep{suadaa2021towards}  leveraged pretrained language model for fact inference in numerical table contents.~%
However, these approaches seldom perceived and integrated the complete table information into the fine-tuning of the pretrained model.
A table-to-text pretrained model~\citep{xing2021structure} was proposed though, the large and diversified table corpus is often unavailable. 
In addition, recent works on fact verification taking tabular as  input~\citep{yin2020tabert,dong2021structural} have suggested the effectiveness of the table-structure-aware pretrained model.

\vspace{-1ex}
\subsection{Text Deliberation}
\vspace{-.5ex}

The encoder-decoder framework has been widely applied to neural machine translation, while the subsequent words are often invisible on the target side when decoding a sequence.~%
To alleviate this, researchers proposed to decode and refine the output sequence in multiple passes, like human cognitive behavior when polishing an article.~%
Studies have been made on text deliberation, 
such as the solution with two separate stages (i.e., generating and polishing)~\citep{niehues2016pre}, combining two separate stages as one framework~\citep{xia2017deliberation}, and deliberating generated text in multiple passes adaptively via reinforcement learning~\citep{geng2018adaptive} or customized evaluating architecture~\citep{li2021rewriter}.~%
To the best of our knowledge, we are the first to apply the deliberation mechanism to the table-to-text problem.

\vspace{-.5ex}
\section{Preliminaries}


\vspace{-1ex}
\subsection{Problem Formulation}
\vspace{-.5ex}

Our table-to-text problem takes a table as input, and we formulate a table as a sequence of records: $ T = \{ \tau_{1, 1}, \tau_{1, 2}, \cdots, \tau_{i, j}, \cdots, \tau_{m, n} \} $, where $m$ and $n$ denote the number of rows and columns of $T$, respectively.
Then, we aim to generate a document $ Y $ containing words $ Y = y_1 y_2 \cdots y_l $ that can describe the content of $T$ precisely, where $l$ is the document length.
Formally, given a table $T$, the table-to-text model is excepted to generate a descriptive document $Y$ in an auto-regressive way

\vspace{-4ex}

\begin{equation*}
y_i = \argmax  P( y_{i} \mid T, y_{1} y_{2} \cdots y_{i-1} ; \, \Theta), \, i = 1, \cdots, l 
\end{equation*}

\vspace{-1.5ex}

\noindent
where $\Theta$ is the set of model parameters.

\subsection{Data}

{\textbf{NumericNLG Dataset}.}
The numericNLG dataset was released by~\citep{suadaa2021towards}. 
In this dataset, the tables demonstrate experimental results in research papers, 
thus, most of the table contents are numerical values.
We use this dataset to evaluate the accuracy and smoothness of the generated descriptions for the table with numerical content.
In particular, for each table of numericNLG, $\texttt{<table\_id>}$ acts as the pronoun of the table, and $\texttt{<caption>}$ is the descriptive text of the table. 
Moreover, for each cell of a table, there are $\texttt{<metric>}$, $(\texttt{row and column})$ $\texttt{<header>}$, and $\texttt{<value>}$ as different views of a cell.


\noindent{\textbf{Totto Dataset.}}~%
The Totto dataset~\citep{parikh2020totto} is an open-domain table-to-text dataset collected from Wikipedia.~%
The table contents are mainly in text form.~%
The metadata of the Totto dataset includes $\texttt{<page\_title>}$, 
$\texttt{<section$ $\_title>}$ 
and $\texttt{<section\_text>}$.~%
In detail, each cell of a table has corresponding $\texttt{<header>}$ and $\texttt{<value>}$.~%
Unlike numericNLG, textual content in our Totto dataset accounts for 62.4\%, which can evaluate the text generation effectiveness for the tables with textual records.




\begin{figure}[t]
    \centering
    \includegraphics[width=0.98\linewidth]{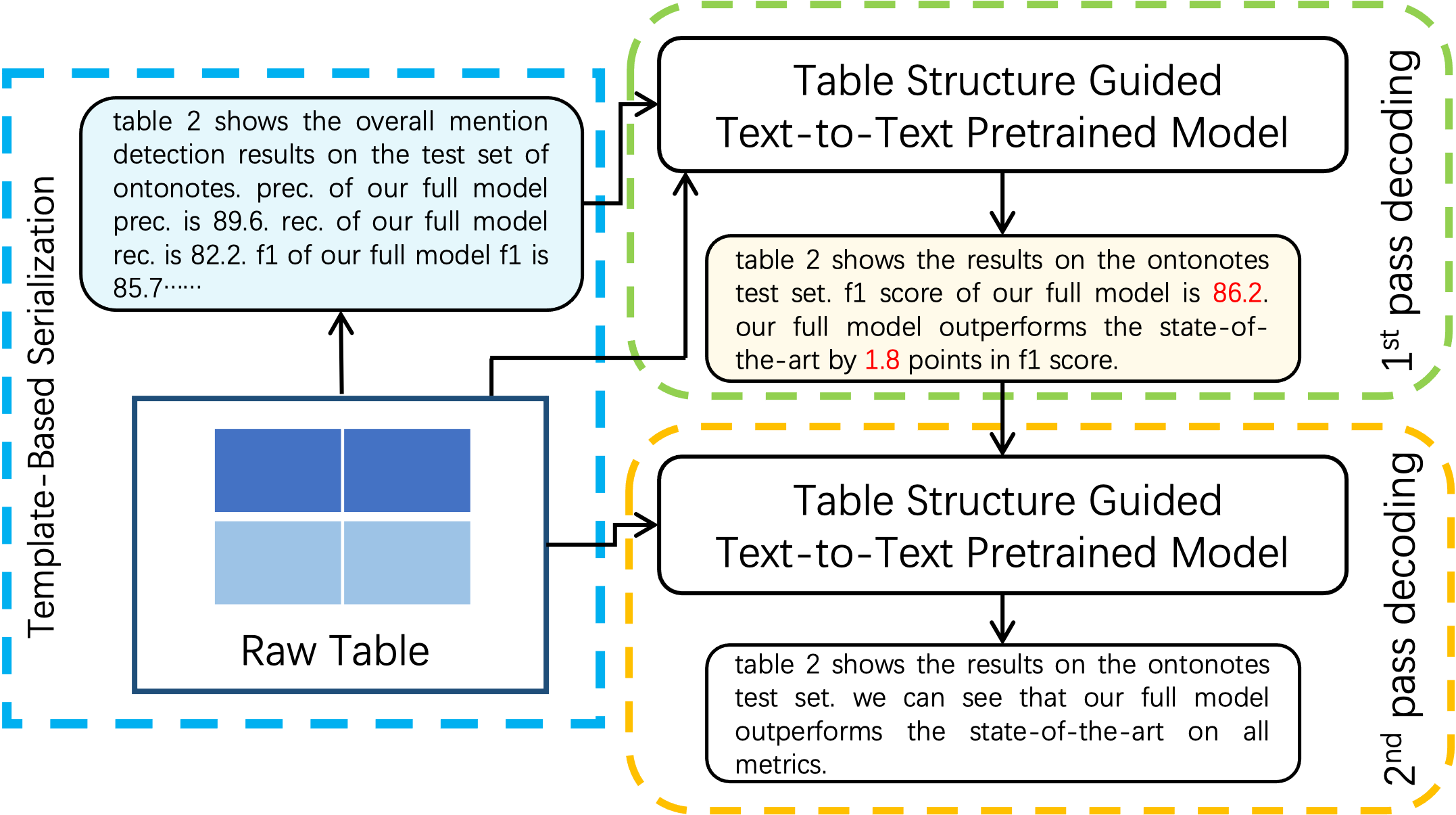}
    \vspace{-1.5ex}
    \caption{
    The framework overview of {\ourframework}.
    }
    \label{fig:overview} 
    \vspace{-2.5ex}
\end{figure}

\section{Methodology}


In this section, we introduce the proposed framework in detail.~%
As shown in \cref{fig:overview}, our framework mainly consists of three components, i.e., \emph{template-based table serialization}, \emph{table-structure-aware fine-tuning}, and \emph{text deliberation}. 
Specifically, we first produce a sequence describing the table contents with customized templates.~%
The templates we adopted do not require the target cells to be labeled. 
Then, to generate informative text, we adopt full table representation learning to guide the description generation, such that the outcome text is capable of emphasizing and delineating the facts in the table from a macroscopic perspective.~%
Finally, we employ and adapt the multi-pass decoder to our data-to-text problem, which can further fine-tune the generated table description. 
Technical details for all three modules will be introduced separately in the following subsections.

\subsection{Template-based Table Serialization}

To well harness the expressive power of the text-to-text pretrained model for the input table, it is necessary to serialize the raw  table first. 
The template-based representation offers us a simple yet effective linearization approach to generating descriptive texts which can reflect the facts in a table without yielding an intractable downstream model.

In particular, the templates we adopted in this work are devised to mention all the available facts in the table without knowing the emphasized cells in advance, which is different from~\citep{suadaa2021towards}. 
The template for describing facts consists of two parts:

\vspace{-1ex}
\begin{enumerate}
    \item The title or descriptive text that comes with the table.
    
    \vspace{-1.5ex}
    \item A series of expressions, in which each one describes the content of a cell.
\end{enumerate}
\vspace{-1ex}

More specifically, for the numericNLG dataset, we apply the following template:  

\vspace{-1ex}
\begin{itemize}
    \item[] 
    $\texttt{<table\_id>}$ shows $\texttt{<caption>}$.
    $\texttt{<metric}_{1,1}$ $\texttt{>}$ of $\texttt{<header}_{1,1}\texttt{>}$ is  $\texttt{<value}_{1,1}$$\texttt{>}$, 
    $\cdots$, 
    $\texttt{<me}$ $\texttt{tric}_{i,j}$$\texttt{>} $ of 
    $\texttt{<header}_{i,j}$$\texttt{>}$ is  $\texttt{<value}_{i,j}\texttt{>} $, 
    $\cdots$.
\end{itemize}
\vspace{-.5ex}

\noindent
For the Totto dataset, we apply another template: 

\vspace{-.5ex}
\begin{itemize}
    \item[]
    As $\texttt{<page\_title>}$ $\texttt{<section\_title>}$, 
    $\texttt{<se}$ $\texttt{ction\_text>}$.
    $\texttt{<header}_{1,1}\texttt{>}$ is $\texttt{<value}_{1,1}\texttt{>} $,
    $\cdots$, 
    $\texttt{<hea}$$\texttt{der}_{i,j}\texttt{>}$ is $\texttt{<value}_{i,j}\texttt{>} $,
    $\cdots$ .
\end{itemize}
\vspace{-.5ex}

\noindent
The second part of the template enumerates all the cells in the table. 
This preliminary table representation, denoted by $T_{S}$, covers all the available facts in a raw table. 
Note that, the templates we adopt may encounter the content selection problem. 
In table-to-text applications, target cells in the input table are often not highlighted and the generated table description should emphasize certain cells.

\subsection{Table-Structure-Aware Text Generation}

A text-to-text pretrained model can take the large-scale corpus as input to possess vast knowledge and generate texts in an unsupervised way so that it has been widely applied to text-generation tasks. 
When handling a specific text generation task, it is effective to fine-tune the pretrained model on new data. 
However, for the table-to-text task, some hidden information, like table structure, is most likely to be overlooked, though the drafted $T_{S}$ mentions all the available facts in the table. 
Thus, we propose to exploit table structure information to guide fine-tuning of the text-to-text pretrained model.

\begin{figure}[t]
    \centering
    \includegraphics[width=0.98\linewidth]{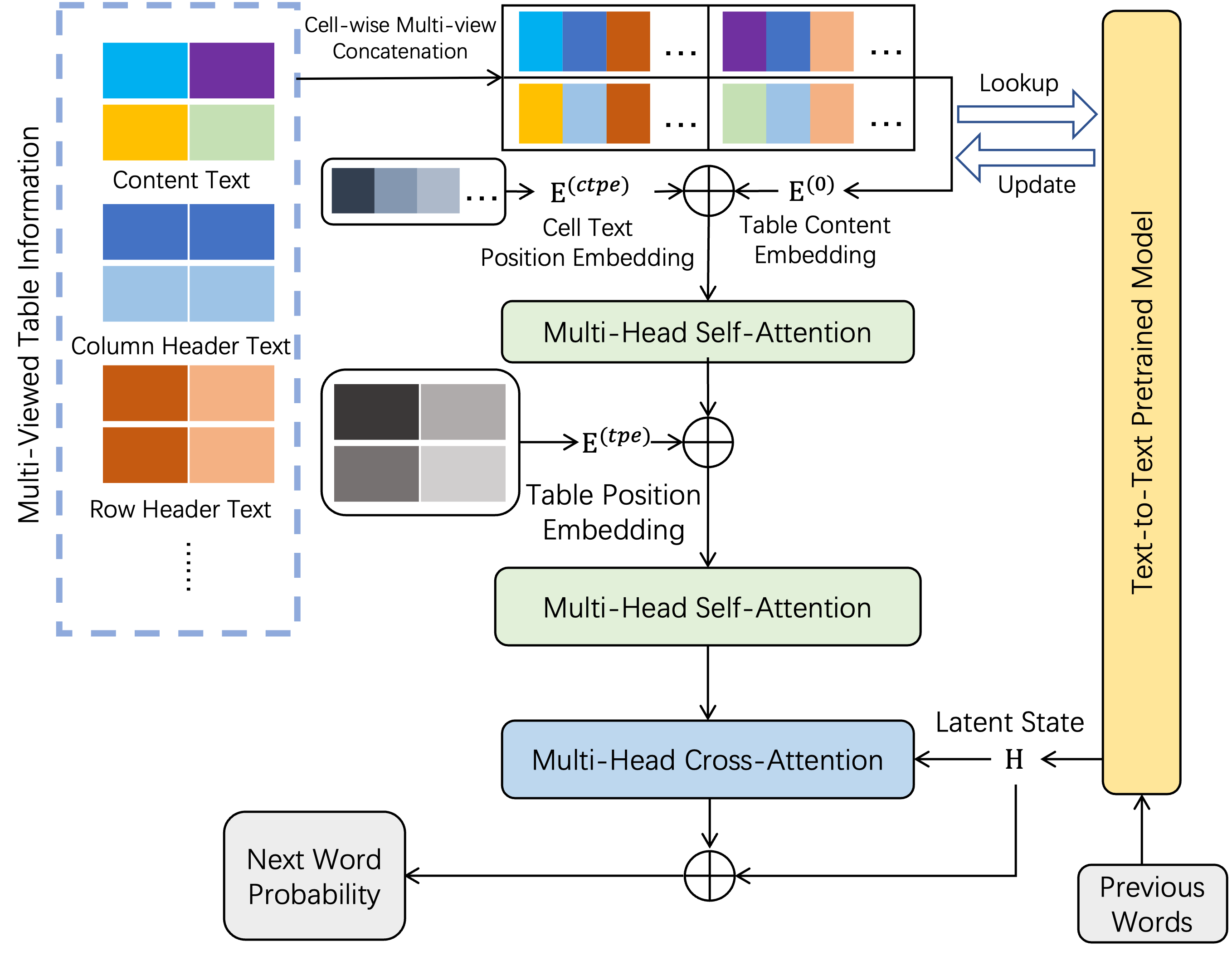}
    \caption{
    The architecture of table-structure-aware text generation model (i.e., \ourmodel).
    }
    \vspace{-2ex}
    \label{fig:table_represent_model}
\end{figure}

As shown in~\cref{fig:table_represent_model}, we first encode the table content in a multi-view fashion. 
To be specific, given a cell $\tau_{i, j}$ in a table $T$, it can be viewed from different perspectives, such as the value of $\tau_{i, j}$, the row header of $\tau_{i, j}$, and the column header of $\tau_{i, j}$, etc.
Then, we treat the $k$-th view of $\tau_{i, j}$ as a token sequence which is denoted by $x^{(k)}_{i, j}$.
Afterward, we pad $x^{(k)}_{i, j}$ with placeholders (if necessary) and concatenate these token sequences as follows:
\begin{equation}    \label{eq:multi-viewed-token-seq}
    \mathbf{x}_{i, j} =  x^{(1)}_{i, j} \, \circledast \, x^{(2)}_{i, j} \, \circledast \, \cdots,
\end{equation}
where $ \circledast  $ denotes the concatenation operator, 
and the multi-viewed representation of a table $T$ is denoted as $ \mathbf{X} = [ \mathbf{x}_{1, 1},   \cdots, \mathbf{x}_{i, j}, \cdots ,  \mathbf{x}_{m, n} ]  $. 
Each token of $\mathbf{x}_{i, j}^{(k)}$ can be encoded as a $d$-dimensional embedding by looking up the text-to-text pretrained model and updated accordingly when fine-tuning the pretrained model.
In this way, we can obtain the semantic representation of  table $T$, which is denoted by $ \mathrm{E}^{(0)} \in \mathbb{R}^{m \times n \times s \times d}$, where $s$ is the length of concatenated sequence $ \mathbf{x}_{i, j} $.

To realize \ourmodel for table-to-text, we propose to employ multi-head attention~\citep{vaswani2017attention} to guide fine-tuning of the text-to-text pretrained model. 
In particular, we adopt three  multi-head attention (MHA) layers  to interactively extract the information in the table in a hierarchical way. 
Specifically, the MHA layer is defined as:  
%
\begin{equation*}
    \begin{aligned}
        &   \,
            Q_i = Q W_{i}^{Q}, \, K_i = K W_{i}^{K}, \, V_i = V W_{i}^{V}   \\
        &   \text { head }_{i} = \text { Attention }(Q_i, K_i, V_i) = \operatorname{softmax}\left(\frac{Q_i     K_i^{\top}}{\sqrt{d}}\right) V_i, \\
        &
        \operatorname{MHA}({Q}, {K}, {V}) = \left[\text{ head }_{1} , \cdots, \text { head }_{{h}} \right] {W}^{O},
    \end{aligned}
\end{equation*}
%
where ${Q}$, ${K}$, ${V}$ represent the query, key and value in the attention mechanism, respectively.


\begin{figure*}[t]
    \centering
    \begin{subfigure}[t]{0.375\textwidth}
      \includegraphics[width=\textwidth]{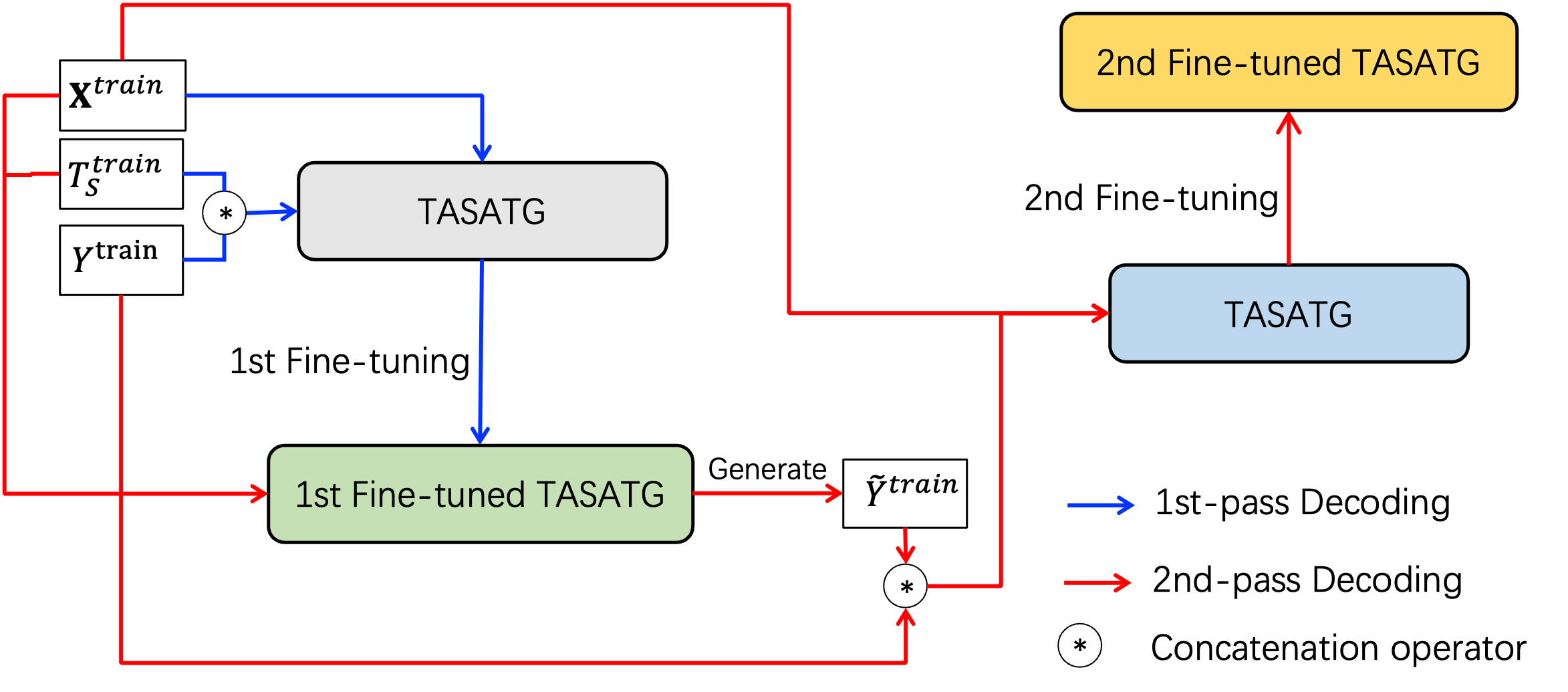}
      \caption{Training.} 
      \label{fig:process:train}
    \end{subfigure}
    \hspace{1ex}
    \begin{subfigure}[t]{0.43\textwidth}
      \includegraphics[width=\textwidth]{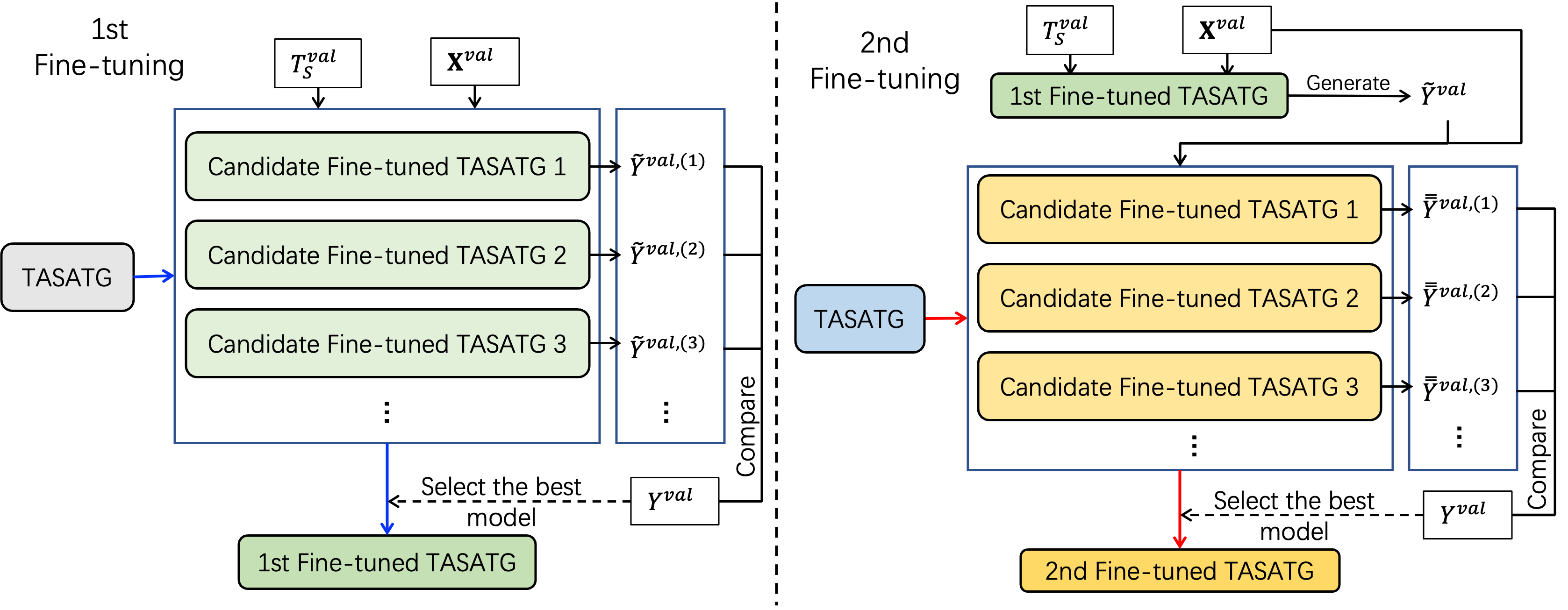}
      \caption{First and second fine-tuning of \ourmodel~with validation data.}
      \label{fig:process:validate}
    \end{subfigure} 
    \hspace{1ex}
    \begin{subfigure}[t]{0.13\textwidth}
      \includegraphics[width=\textwidth]{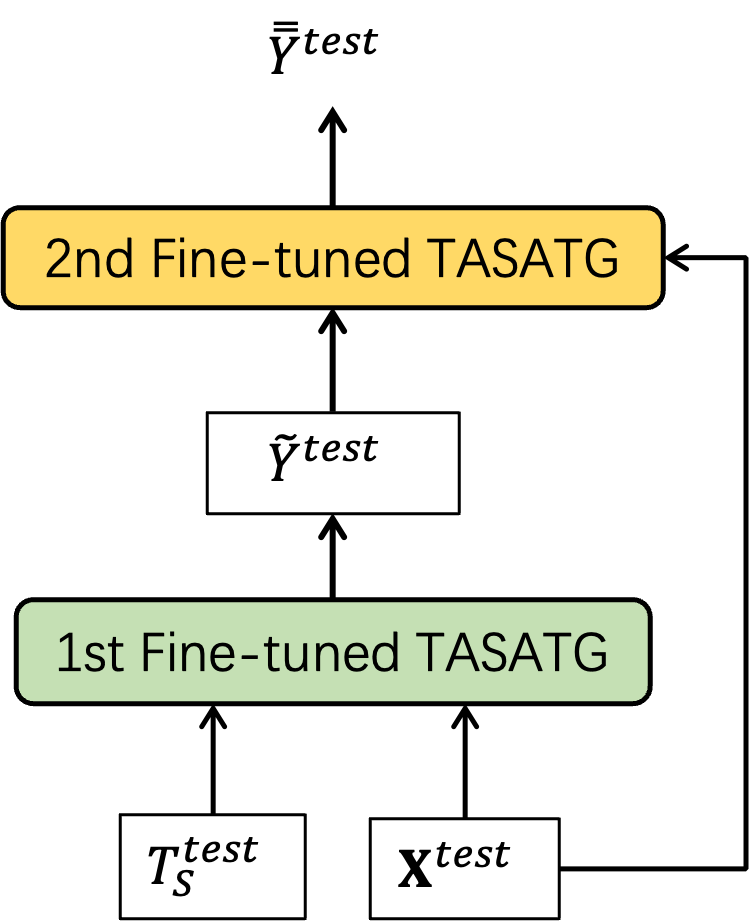}
      \caption{Testing.} 
      \label{fig:process:test}
    \end{subfigure}
    \vspace{-.75ex}
    \caption{
    Training, validation and testing procedures of the proposed \ourframework~approach.
    }
    \vspace{-1ex}
    \label{fig:process}
\end{figure*}


As illustrated in \cref{fig:table_represent_model}, in the first MHA layer, we add a cell text position embedding ($ \mathrm{E}^{(ctpe)} \in \mathbb{R}^{s \times d}$) to each cell of the aforementioned $ \mathrm{E}^{(0)} $, and feed it to the multi-head attention to implement cell text self-attention, 
\begin{equation}    \label{eq:e1}
\begin{aligned}
    \widetilde{\mathrm{E}_{0}}  &= \mathrm{E}^{(0)} \oplus \mathrm{E}^{(ctpe)}, \\
    \mathrm{E}^{(1)}            &= \operatorname{MHA}(\widetilde{\mathrm{E}^{(0)}}, \widetilde{\mathrm{E}^{(0)}}, \widetilde{\mathrm{E}^{(0)}}) ,\\
    \mathrm{E}^{(1)}            &= \frac{1}{s} \sum_{i=1}^{s} (\mathrm{E}^{(1)} [:, :, i, :]) \, ,
\end{aligned}
\end{equation}
%
where $ \oplus $ denotes the element-wise addition operation. 
Consequently, $\mathrm{E}^{(1)} \in \mathbb{R}^{m \times n \times d} $ can be deemed as an initial aggregated table  representation. 
Next, in the second MHA layer, we add a table position embedding ($ \mathrm{E}^{(tpe)} \in \mathbb{R}^{m \times n \times d} $) to $ \mathrm{E}^{(1)} $ to implement table structure self-attention,  
\begin{equation}    \label{eq:e2}
\begin{aligned}
    & \widetilde{\mathrm{E}^{(1)}} = \mathrm{E}^{(1)} \oplus \mathrm{E}^{(tpe)}, \\
    & \mathrm{E}^{(2)} = \operatorname{MHA}(\widetilde{\mathrm{E}^{(1)}}, \widetilde{\mathrm{E}^{(1)}}, \widetilde{\mathrm{E}^{(1)}}) .
\end{aligned}
\end{equation}
%
$\mathrm{E}^{(2)}\in \mathbb{R}^{m\times n \times d}$ is the table-structure-aware representation. 
Moreover, in the third MHA layer, we apply a multi-head cross-attention to take the hidden state of the text-to-text pretrained model (denoted by $\mathrm{H} \in \mathbb{R}^{s \times d} $)  as the attention query, such that we can focus on the important cells of the table, 
\begin{equation}    \label{eq:outcome-embed}
    \widetilde{\mathrm{H}} = \operatorname{MHA}(\mathrm{H}, \mathrm{E}^{(2)}, \mathrm{E}^{(2)}) \oplus \mathrm{H}. 
\end{equation}
%
This new hidden state $\widetilde{\mathrm{H}}$ guided by the table representation will replace the original hidden state $\mathrm{H}$ in the text-to-text pretrained model to generate the probability of the next word. 

Note that, the cross attention weights on different table cells based on the previous words can realize the content selection automatically. 
In addition, we implement the text-to-text pretrained model with GPT2~\citep{radford2019language}, which adopts a decoder-only Transformer architecture.

\subsection{Text Deliberation}

The encoder-decoder framework applied in many sequence generation tasks often adopts a one-pass process while decoding a sequence.~%
Though efficient, the one-pass decoder cannot perceive future context for further text deliberation.~%
Multi-pass decoder extends the capability of generating more refined text by exploring global information in the sequence~\citep{niehues2016pre,xia2017deliberation}. 

For the text-to-text pretrained model, due to the huge amount of parameters of the pretrained language model, it is unwise to directly combine the models in different passes.~%
A common solution is to concatenate the original serialized table content and the text generated in the previous pass to fine-tune the pretrained model in the next-pass decoding.
However, in this way, the length of input text probably exceeds the limit of the text-to-text pretrained model, and the time complexity is too high.

To effectively implement the fine-tuning of the text-to-text pretrained model in multiple passes, as shown in \cref{fig:process:train,fig:process:validate}, we take the table representation as the ``original text'' and feed the text generated in the first-pass fine-tuning plus the table representation to the second-pass fine-tuning. 
Note that, as shown in \cref{fig:process:train}, we separately fine-tune the table-to-text generation task and the text-to-text deliberation task with two independent \ourmodel ~models, and each of them takes a text-to-text pretrained model as the backbone.

\section{Experiments}


\begin{table*}[t]
\centering
\tabcolsep 0.06in
\caption{
Performance comparisons of the automatic evaluation on the numericNLG dataset.
}
\label{table:automatic-eval:numericNLG}
\vspace{-1ex}
\begin{tabular}{l c  c  c  c  c  c  c}
\toprule
Method              & BLEU-1 &BLEU-2 &BLEU-3 &BLEU-4 &METEOR &ROUGE-L \\
\midrule
Template-based Method 
                    & 10.28& 5.52& 2.83& 1.14& 11.31& 11.49      \\
Pointer Generator 
                    & $5.10_{\pm0.59}$& $2.71_{\pm0.19}$& $1.16_{\pm0.17}$& $0.56_{\pm0.04}$& $7.82_{\pm0.15}$& $15.21_{\pm0.14}$       \\
TRM 
                    & $14.16_{\pm0.97}$& $6.05_{\pm0.50}$& $2.11_{\pm0.13}$& $0.80_{\pm0.12}$& $9.72_{\pm0.94}$& $12.72_{\pm0.80}$      \\ 
Fine-tuned GPT2     
                    & $16.13_{\pm0.56}$& $9.02_{\pm0.31}$& $4.68_{\pm0.22}$& $2.20_{\pm0.22}$& $10.14_{\pm0.32}$& $17.48_{\pm0.36}$       \\
TableGPT
                    & $18.69_{\pm0.39}$& $8.21_{\pm0.24}$ & $3.31_{\pm0.19}$& $1.51_{\pm0.14}$& $11.06_{\pm0.18}$& $16.90_{\pm0.27}$    \\
\midrule
\ourframework$_{w/o \,\, \text{TAS}} $
                    & $18.20_{\pm2.40}$& $9.74_{\pm1.01}$& $4.38_{\pm0.31}$& $1.98_{\pm0.39}$& $10.64_{\pm0.86}$& $19.29_{\pm1.77}$ \\
\ourframework$_{w/o \,\, \text{D}}$
                    & $18.02_{\pm0.50}$& $10.06_{\pm0.25}$& $\textbf{5.20}_{\pm0.13}$& $\textbf{2.47}_{\pm0.20}$& $10.99_{\pm0.29}$& $18.57_{\pm0.27}$      \\
\ourframework$_{w/o \,\, 1^{\text{st}}\text{-TAS}}$
                    & $20.07_{\pm1.94}$& $10.35_{\pm0.69}$& $4.67_{\pm0.35}$& $2.05_{\pm0.34}$& $11.52_{\pm0.80}$& $20.10_{\pm0.62}$      \\
\ourframework
                    & $\textbf{21.81}_{\pm1.13}$ & $\textbf{11.03}_{\pm0.11}$ & $4.92_{\pm0.22}$ & $2.15_{\pm0.39}$ & $\textbf{11.87}_{\pm0.40}$ & $\textbf{20.40}_{\pm0.80}$      \\
\bottomrule
\end{tabular}
\end{table*}

\begin{table*}[t]
\centering
\tabcolsep 0.06in
\caption{
Performance comparisons of the automatic evaluation on the Totto dataset.
} 
\label{table:automatic-eval:totto}
\vspace{-1ex}
\begin{tabular}{ l c  c  c  c  c  c  c }
\toprule
Method              & BLEU-1 &BLEU-2 &BLEU-3  & BLEU-4 &METEOR &ROUGE-L \\
\midrule
Template-based Method
                    & 0.84& 0.43& 0.23& 0.09& 4.59& 1.51      \\
Pointer Generator                                               
                    & $11.34_{\pm1.57}$& $2.05_{\pm0.83}$& $0.45_{\pm0.27}$& $0.35_{\pm0.13}$& $5.38_{\pm0.78}$& $14.46_{\pm1.46}$        \\
TRM
                    & $10.21_{\pm1.79}$ & $3.44_{\pm0.88}$  & $1.21_{\pm0.48}$ & $0.54_{\pm0.25}$& $9.30_{\pm1.16}$ & $11.52_{\pm2.03}$        \\
Fine-tuned GPT2     
                    & $9.53_{\pm0.51}$  & $3.65_{\pm0.34}$  & $1.18_{\pm0.37}$& $0.40_{\pm0.26}$& $9.89_{\pm0.39}$& $10.69_{\pm0.27}$        \\
TableGPT 
                    & $6.80_{\pm0.26}$ & $3.51_{\pm0.22}$ & $1.33_{\pm0.21}$ & $0.76_{\pm0.12}$ & $11.10_{\pm0.42}$& $11.73_{\pm0.44}$       \\ 
\midrule
\ourframework$_{w/o \,\, \text{TAS}} $
                    & $13.70_{\pm0.90}$ & $4.44_{\pm0.69}$  & $1.28_{\pm0.47}$& ${0.65}_{\pm0.35}$& $10.79_{\pm0.83}$& $14.47_{\pm1.11}$       \\
\ourframework$_{w/o \,\, \text{D}}$
                    & $10.03_{\pm0.39}$ & $4.42_{\pm0.29}$  & $1.64_{\pm0.36}$ & $0.71_{\pm0.38}$& $10.29_{\pm0.49}$& $10.67_{\pm0.34}$        \\
\ourframework$_{w/o \,\, 1^{\text{st}}\text{-TAS}}$
                    & $13.90_{\pm0.60}$ & $5.07_{\pm0.61}$  & $1.68_{\pm0.52}$ & $\textbf{0.79}_{\pm0.25}$ & $10.98_{\pm0.40}$ & $14.88_{\pm0.71}$  \\
\ourframework
                    & $\textbf{14.19}_{\pm1.08}$ & $\textbf{5.17}_{\pm0.38}$ & $\textbf{1.71}_{\pm0.32}$ & $0.78_{\pm0.21}$ & $\textbf{11.65}_{\pm0.71}$ & $\textbf{14.96}_{\pm1.10}$ \\
\bottomrule
\end{tabular}
\end{table*}

\subsection{Experimental Settings}

{\bf Data.}~%
We conducted experiments on the aforementioned datasets, i.e., numericNLG and Totto.
The statistics of the numericNLG dataset can be found in ~\citep{suadaa2021towards}. 
Besides, the size of the original Totto dataset is 120K, which is much larger than the numericNLG dataset. 
To evaluate different methods for table-to-text with comparable data size, for the Totto dataset, we filtered out the tables with fewer rows and columns, i.e., \#rows $<$ 8 and \#columns $<8$, such that the filtered Totto dataset contains 1.8K tables.
Then, we randomly selected 1.2K{\footnote{The size of numericNLG data is 1.3K.}} tables to generate the new Totto dataset.


\noindent
{\bf Evaluation Metrics.}~%
We calculated BLEU (from gram-1 to gram-4)~\citep{papineni2002bleu}, ROUGE-L~\citep{lin2004rouge} and METEOR~\citep{denkowski2014meteor} to evaluate the quality of the generated text. 
The BLEU-n with a small value of n measures the accuracy of the word level, and the BLEU-n with a large n can measure the fluency of the sentence. The ROUGE-L measures the recall rate based on the longest common sequence between source and target texts. The METEOR is based on the harmonic mean of unigram precision and recall, with recall weighted higher than precision.
These metrics are widely used to measure the accuracy and fluency of the generated sentence.

\noindent
{\bf{Baselines.}}
We compare \ourframework with the following baselines. 

\vspace{-1ex}
\begin{itemize}
    \item 
    {\bf Template-based Table Serialization}. 
    We use the template designed for table serialization as a baseline. 
    Note that, the token sequence generated by the template-based method is denoted as $T_{S}$.
    
    \vspace{-1ex}
    \item 
    {\bf Pointer Generator}~\citep{see2017get}. 
    This is a seq2seq model with the attention and copy mechanism. 
    We take $T_{S}$ as input for the pointer generator model. 
    
    \vspace{-1ex}
    \item 
    {\bf TRM}. 
    We implemented a simplified version of the proposed \ourframework that omits the possessed knowledge in the pretrained language model and removes text deliberation for focusing on {\underline{t}}able {\underline{r}}epresentation {\underline{m}}odeling, namely TRM. 
    In particular, TRM adopts the architecture of GPT2 but initializes the parameters randomly and trains 100 epochs at most  for fine-tuning.
    Besides, TRM takes $T_{S}$ plus the table structure representation as input and is fed with $T_S$ in the inference phase.
    
    \vspace{-1ex}
    \item 
    {\bf Fine-tuned GPT2}~\citep{radford2019language}.~%
    We take the concatenation of $T_{S}$ and $Y$ as the input for fine-tuning. 
    In the inference phase, we only feed $T_{S}$ to the model to generate $Y$ starting after the last token of $T_{S}$. 
    
    \vspace{-1ex}
    \item 
    {\bf TableGPT}~\citep{gong2020tablegpt}.~%
    TableGPT is a state-of-the-art table-to-text method.
    To improve the text fidelity and exploit the structural information at the same time,
    TableGPT employs a multi-task learning paradigm consisting of two auxiliary tasks, that is,  one task reconstructs the table structure from representations of GPT2, and the other aligns the tables and the information in the generated text. 
    \vspace{-1ex}
\end{itemize}

\noindent
{\bf{Implementation Details.}~}%
The split settings for training, validation and, testing were 1084:136:135 {\footnote{This setting follows the experiments of~\citep{suadaa2021towards}.}}  for the numericNLG dataset and 960:120:120 for the Totto dataset, respectively. 
Regarding automatic evaluation, all results of deep models were obtained by conducting experiments on a Linux machine with Nvidia A100 GPU, and the averaged results of 5 runs were reported. 
Besides, an Adam optimizer was utilized (with an initial learning rate of 3e-5) for GPT2 fine-tuning, and the training was iterated in 20 epochs at most. 
A beam search algorithm was adopted when decoding a sequence and the beam width was set to 5 {\footnote{Our implementation is available at \url{https://github.com/ramber1836/TASD}.}}.

\subsection{Automatic Evaluation}

The comparisons of automatic evaluation results between \ourframework and other baselines can be found in \cref{table:automatic-eval:numericNLG,table:automatic-eval:totto}.
In general, \ourframework outperforms the baselines for all the metrics on two datasets. 
In particular, compared to the reported best result of all the baselines, \ourframework achieves improvements of 3.12 for BLEU-1 (18.69 $\rightarrow$ 21.81), 2.01 for BLEU-2 (9.02 $\rightarrow$ 11.03), 0.24 for BLEU-3 (4.68 $\rightarrow$ 4.92), 0.56 for METEOR (11.31 $\rightarrow$ 11.87), and 2.92 for ROUGE-L (17.48 $\rightarrow$ 20.40) on the numericNLG dataset, 
and 2.85 for BLEU-1 (11.34 $\rightarrow$ 14.19), 1.52 for BLEU-2 (3.65 $\rightarrow$ 5.17), 0.38 for BLEU-3 (1.33 $\rightarrow$ 1.71), 0.02 for BLEU-4 (0.76 $\rightarrow$ 0.78), 0.55 for METEOR (11.10 $\rightarrow$ 11.65), and 0.50 for ROUGE-L (14.46 $\rightarrow$ 14.96) on the Totto dataset. 
In other words, for different types of source tables, \ourframework generates better descriptive texts w.r.t. accuracy at the word level, recall of the sequence, and fluency of sentences.

Besides, we have the following observations: 
{\bf 1)} 
The template-based method performs much better on the numericNLG dataset compared to the Totto dataset, since the referenced table descriptions in numericNLG  were collected from scientific papers, however, the table summaries in the Totto dataset are more diverse.
{\bf 2)} 
In the Totto dadaset, the pointer generator model tends to cover more words in descriptive text and generate more fluent sentences than the template-based method,  as the contents in source tables of the Totto dataset are mostly linguistic. 
This can also explain why the pointer generator performs worse than the template-based method on the numericNLG dataset  w.r.t. BLEU and METEOR.  
{\bf 3)}
Fine-tuned GPT2 can generate more faithful and fluent text than other baselines  (refer to \cref{table:automatic-eval:numericNLG,table:automatic-eval:totto}) most of the time, which validates the effectiveness of the pretrained language model.
{\bf 4)}
In general, TableGPT performs better, and even the best, among all the baselines. 
In the numericNLG dataset, the headers of the input tables (a.k.a. the attributes of records for TableGPT) are more diverse, which may explain why the performance of TableGPT is not promising as expected on the numericNLG dataset.
{\bf 5)}
TRM can generate comparable, or even better descriptive text as fined-tuned GPT2, which further suggests the effectiveness of table structure understanding.

\begin{figure*}
    \centering
    \includegraphics[width=1.0\linewidth]{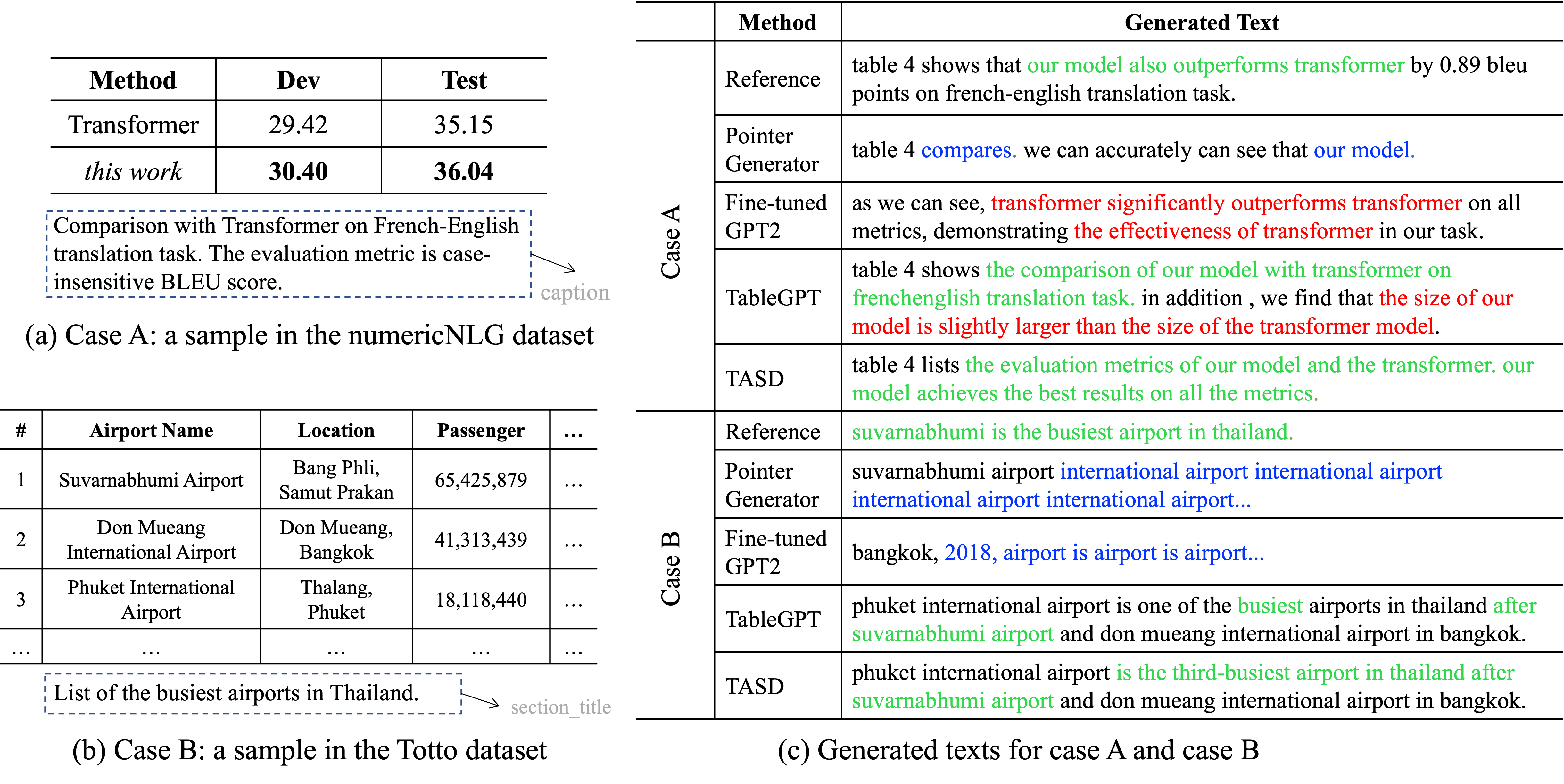}
    \vspace{-3ex}
    \caption{
    Two examples of the generated table descriptions.
    }
    \label{fig:qualitative-analysis}
    \vspace{-1ex}
\end{figure*}

\subsection{Ablation Analysis}

Moreover, to verify the effectiveness of different modules, we compare TASD with its variants.

\begin{itemize}
    \item 
    After generating text with fine-tuned GPT2, we fed the generated text concatenated with $T_{S}$ to another fine-tuned GPT2 to realize the second-pass decoder without table structure representation.  
    
    \vspace{-1ex}
    \item 
    We implemented \ourframework without deliberating on the outcome text, which means that we realized \ourmodel based on GPT2 in a one-pass forward process. 
    
    \vspace{-1ex}
    \item 
    \ourframework$_{w/o \,\, 1^{\text{st}}\text{-TAS}}$. 
    We removed table structure modeling in the first-pass decoding from \ourframework, which was implemented by taking the fine-tuned GPT2 as the first-pass decoder and the table-structure-aware fine-tuned GPT2 as the second-pass decoder.

\end{itemize}

As can be seen in \cref{table:automatic-eval:numericNLG,table:automatic-eval:totto}, \ourframework$_{w/o \,\, \text{TAS}} $ performs worse than \ourframework under all metrics, since the table structure modeling can benefit the fine-tuning of GPT2. 
This can also be validated by comparing fine-tuned GPT2 to \ourframework$_{w/o \,\, \text{D}}$. 
Besides, the effectiveness of deliberating text can be proven by comparing \ourframework$_{w/o \,\, \text{D}}$ to \ourframework (this can also be validated by comparing fine-tuned GPT2 to \ourframework$_{w/o \,\, \text{TAS}}$). 
While text deliberation may harm sentence fluency as depicted by the results of these methods w.r.t. BLEU-3 \& 4 in \cref{table:automatic-eval:numericNLG}. 
In addition, \ourframework$_{w/o \,\, 1^{\text{st}}\text{-TAS}}$ outperforms  \ourframework$_{w/o \,\, \text{TAS}} $ under all metrics suggesting that taking the table representation as the ``original text'' in the deliberation mechanism is also effective.

\subsection{Qualitative Analysis}

Figs. \ref{fig:qualitative-analysis}(a) and (b) show two selected source tables and corresponding descriptive texts (i.e., caption and section\_text) in numericNLG and Totto datasets.
\cref{fig:qualitative-analysis}(c) demonstrates the generated descriptions by different methods. 
The text that correctly reflects the facts of the source table is in green, the erroneous text is in red, and the confusing text is in blue. 
We can see that,  there are many grammatical errors in the text produced by the pointer generator. 
Fine-tuned GPT2 tends to repeat phrases and sentences due to the limited knowledge about the input table, which can also explain why the fine-tuned GPT2 can obtain a false high score in BLEU-n as n grows.
Thanks to the semantic knowledge brought by pretraining, fine-tuned GPT2 can generate more natural descriptions, in which, however, perplexing factual errors exist. 
Compared to fine-tuned GPT2, the description generated by \ourframework is more relevant to the table contents. 
Since the target cells are not known in advance, the generated text may miss the emphasized points described in the reference. 
The text generated by TableGPT is also fluent, though counterfactual descriptions may exist.

\begin{table}[t]
\small
\centering
\tabcolsep 0.025in
\caption{
Result of Human Evaluation
}
\label{table:exp-human-eval}
\vspace{-1ex}

\begin{tabular}{ m{0.12\linewidth} | m{0.25\linewidth} | m{0.16\linewidth} | m{0.19\linewidth} | m{0.19\linewidth} }
\toprule
Dataset     & \multicolumn{1}{c|}{Method}            
                                & Grammar       & Coherence \& Concise      &  Factual perspective
\\  \midrule
\multicolumn{1}{c|}{
\multirow[c]{4}{*}{\rotatebox{90}{\small{numericNLG~}}}
}
            & {\small{Pointer Generator}} 
                                & 3.16$_{\pm0.99}$          & 2.73$_{\pm1.20}$                      & 1.54$_{\pm0.69}$
\\  \cline{2-5}     
            & Fine-tuned GPT2   &3.42$_{\pm0.56}$           & 3.11$_{\pm0.58}$                      & 2.51$_{\pm0.45}$
\\  \cline{2-5}     
            & \ourframework$_{w/o \,\, \text{D}}$ 
                                & 3.72$_{\pm0.61}$          & 3.48$_{\pm0.55}$                      & 2.82$_{\pm0.45}$
\\  \cline{2-5}     
            & \ourframework     & {\bf{4.17}}$_{\pm0.72}$   & {\bf{3.98}}$_{\pm0.64}$               & \textbf{3.15}$_{\pm0.73}$
\\  \midrule
\multicolumn{1}{c|}{
\multirow[c]{4}{*}{\rotatebox{90}{\small{Totto~~~~~}}}
}
            & {\small{Pointer Generator}} 
                                & 2.03$_{\pm0.71}$          & 1.89$_{\pm0.82}$                      & 1.56$_{\pm0.55}$
\\  \cline{2-5}     
            & Fine-tuned GPT2   & 2.60$_{\pm0.55}$          & 2.36$_{\pm0.64}$                      & 1.85$_{\pm0.46}$
\\  \cline{2-5}     
            & \ourframework$_{w/o \,\, \text{D}}$
                                & 2.63$_{\pm0.52}$          & 2.46$_{\pm0.60}$                      & 1.89$_{\pm0.46}$
\\  \cline{2-5}     
            & \ourframework     & \bf{3.4}$_{\pm0.66}$     & \bf{3.18}$_{\pm0.70}$                 & \bf{2.25}$_{\pm0.69}$
\\  \bottomrule
\end{tabular}
\vspace{-1ex}
\end{table}

\subsection{Human Evaluation}

We randomly selected 30 samples from the test set in numericNLG and Totto datasets, respectively, and invited 10 volunteers to evaluate the quality of the outcome text by considering three criteria, i.e., grammar, coherence \& concise, and factual perspective (correct and relevant). 
Each criterion has scores of five degrees, ranging from 1 (the worst) to 5 (the best). 
The averaged scores were reported in \cref{table:exp-human-eval}, which show that \ourframework can generate more readable and coherent texts, and describe more correct facts. 
Moreover, the pretrained models consistently achieve better scores than the pointer generator on grammar and coherence because of the expressive power learned from the large-scale corpus.  
In the Totto dataset, the improvement of the table structure modeling is smaller than that of the polishing mechanism, which is consistent with the automatic evaluation results in \cref{table:automatic-eval:totto}.

\begin{table*}[t]
\centering
\tabcolsep 0.06in
\caption{
The performances of \ourframework w/ and w/o the table reconstruction on the numericNLG dataset.
}
\label{table:disscussion-eval:numericNLG}
\begin{tabular}{l c  c  c  c  c  c  c}
\toprule
Method              & BLEU-1 &BLEU-2 &BLEU-3 &BLEU-4 &METEOR &ROUGE-L \\
\midrule
\ourframework$_{w/o \,\, \text{D}}$
                    & $18.02_{\pm0.50}$& $10.06_{\pm0.25}$& ${5.20}_{\pm0.13}$& ${2.47}_{\pm0.20}$& $10.99_{\pm0.29}$& $18.57_{\pm0.27}$      \\
                    

\ourframework$_{w/o \,\, \text{D}\,\,w/ \,\,\text{TRLoss}}$
                    & $20.56_{\pm0.25}$& $\textbf{11.57}_{\pm0.21}$ & $\textbf{5.90}_{\pm0.23}$& $\textbf{2.98}_{\pm0.17}$& $12.00_{\pm0.48}$& $\textbf{20.50}_{\pm0.39}$    \\

                    
\ourframework$_{w/ \,\,\text{TRLoss}
}$
                    & $19.29_{\pm0.38}$ & $10.12_{\pm0.24}$ & $5.32_{\pm0.25}$ & $2.62_{\pm0.22}$ & $\textbf{12.18}_{\pm0.90}$ & $18.95_{\pm0.69}$  \\
                    
\ourframework$_{w/ \,\,\text{TRLoss in 1st pass}
}$
                    & $18.23_{\pm0.68}$ & $9.39_{\pm0.52}$ & $4.64_{\pm0.26}$ & $2.36_{\pm0.24}$ & $11.51_{\pm0.78}$ & $18.13_{\pm0.45}$  \\
                    
\ourframework$_{w/ \,\,\text{TRLoss in 2nd pass}
}$
                    & $19.38_{\pm2.21}$ & $10.33_{\pm1.34}$ & $5.11_{\pm0.73}$ & $2.40_{\pm0.38}$ & $11.35_{\pm0.92}$ & $18.69_{\pm1.05}$  \\
                    
\ourframework
                    & $\textbf{21.81}_{\pm1.13}$ & $11.03_{\pm0.11}$ & $4.92_{\pm0.22}$ & $2.15_{\pm0.39}$ & $11.87_{\pm0.40}$ & $20.40_{\pm0.80}$      \\
\bottomrule
\end{tabular}
\end{table*}

\begin{table*}[t]
\centering
\tabcolsep 0.06in
\caption{
The performances of \ourframework w/ and w/o the table reconstruction on the Totto dataset.
}
\label{table:disscussion-eval:Totto}
\begin{tabular}{l c  c  c  c  c  c  c}
\toprule
Method              & BLEU-1 &BLEU-2 &BLEU-3 &BLEU-4 &METEOR &ROUGE-L \\
\midrule
\ourframework$_{w/o \,\, \text{D}}$
                    & $10.03_{\pm0.39}$ & $4.42_{\pm0.29}$  & $1.64_{\pm0.36}$ & $0.71_{\pm0.38}$& $10.29_{\pm0.49}$& $10.67_{\pm0.34}$        \\
                    
\ourframework$_{w/o \, \text{D} \,\, w/ \,\text{TRLoss}
}$
                    & $9.94_{\pm0.43}$ & $4.35_{\pm0.31}$ & $1.63_{\pm0.31}$ & $0.75_{\pm0.13}$ & $10.37_{\pm0.22}$ & $10.62_{\pm0.60}$     \\

                    
\ourframework$_{w/ \,\,\text{TRLoss}
}$
                    & $\textbf{14.57}_{\pm0.87}$ & $5.22_{\pm0.42}$ & $1.70_{\pm0.49}$ & $\textbf{0.89}_{\pm0.38}$ & $\textbf{11.79}_{\pm0.77}$ & $\textbf{15.28}_{\pm0.86}$  \\
                    
\ourframework$_{w/ \,\,\text{TRLoss in 1st pass}
}$
                    & $14.00_{\pm0.82}$ & $\textbf{5.31}_{\pm0.27}$ & $\textbf{1.72}_{\pm0.25}$ & $0.75_{\pm0.13}$ & $11.02_{\pm0.77}$ & $14.74_{\pm0.51}$  \\
                    
\ourframework$_{w/ \,\,\text{TRLoss in 2nd pass}
}$
                    & $13.89_{\pm0.58}$ & $4.78_{\pm0.61}$ & $1.47_{\pm0.14}$ & $0.52_{\pm0.20}$ & $11.07_{\pm0.66}$ & $14.73_{\pm0.79}$  \\
\ourframework
                    & $14.19_{\pm1.08}$ & $5.17_{\pm0.38}$ & $1.71_{\pm0.32}$ & $0.78_{\pm0.21}$ & $11.65_{\pm0.71}$ & $14.96_{\pm1.10}$ \\
\bottomrule
\end{tabular}
\end{table*}

\section{Discussion}    \label{sec:discuss}

In our work, we devised a two-pass decoder framework dedicated to the table-to-text task with the help of the table-structure-aware text generation model (i.e., \ourmodel). 
However, the effectiveness of the text deliberation for the table-to-text task should be further explored and integrated into the table-structure-aware modeling in a more harmonic manner. 
To discuss the limitation of the text deliberation of \ourframework, we additionally developed a table content reconstruction loss and integrate it into \ourframework in a multi-task learning fashion. 

\begin{figure}[t]
    \centering
    \includegraphics[width=1.0\linewidth]{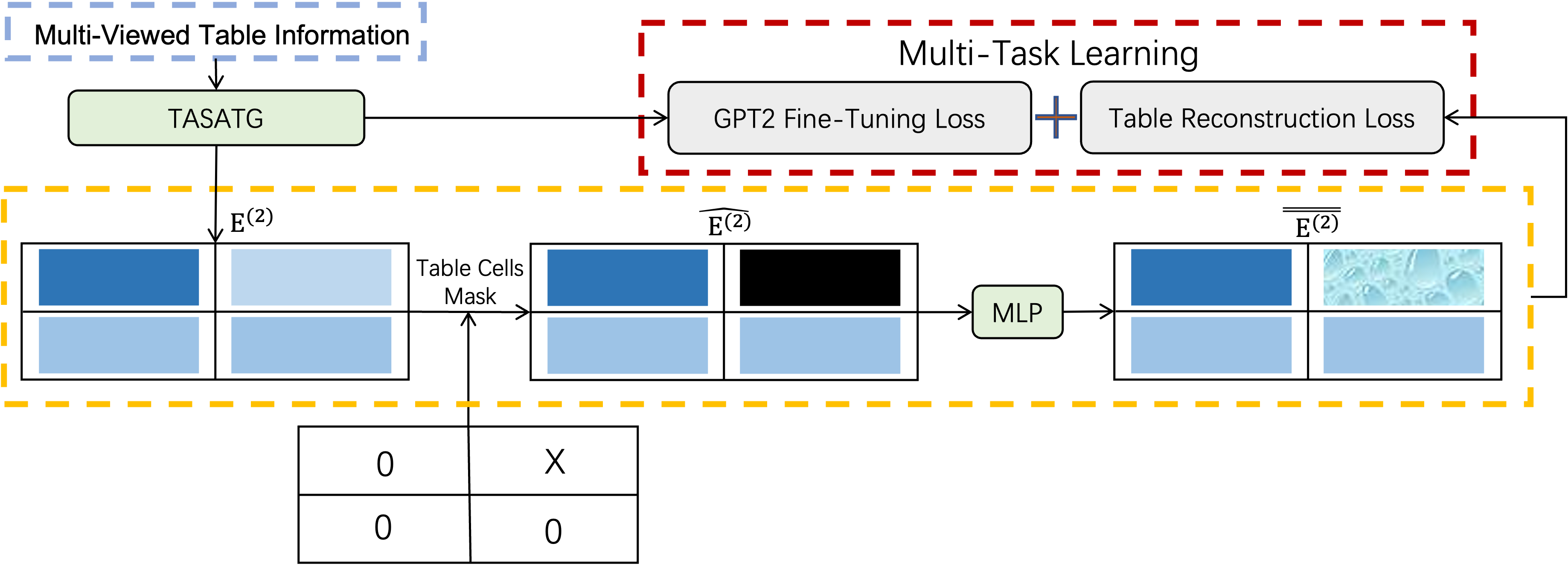}
    \vspace{-2ex}
    \caption{
    Table reconstruction for table-structure-aware modeling enhancement. 
    } \label{fig:mtl_loss}
    \vspace{-2ex}
\end{figure}

Specifically, given the table-structure-aware embedding $\mathrm{E}^{(2)}$ generated with \cref{eq:e2}, we randomly mask certain cells of the input table and yield a partially corrupted embedding of the input table, denoted by $\widehat{\mathrm{E}^{(2)}}$. 
Then, a two-layer MLP (i.e., multi-layer perceptron) is adopted to restore the table-structure-aware embedding. 
Afterward, an MSE (i.e., mean square error) loss is adopted to measure the effectiveness of table reconstruction and further integrated into the \ourframework framework in the multi-task learning paradigm. 
The process of table reconstruction is demonstrated in ~\cref{fig:mtl_loss}.

We carried out a series of experiments to evaluate the performance of \ourframework w/ and w/o the help of table reconstruction loss (i.e., TRLoss) on numericNLG and Totto datasets in terms of BLEU-n (1 to 4), METEOR, and ROUGE-L. 
The results can be found in \cref{table:disscussion-eval:numericNLG,table:disscussion-eval:Totto}. 

According to the results reported on the numericNLG dataset, the TRLoss is helpful in enhancing the capability of table comprehension though, the best performance is achieved by \ourframework$_{w/o\,\text{D}\,\,w/ \, \text{TRLoss}}$.~%
It seems that the performance improvement gained by the table comprehension enhancement is sacrificed after the text deliberation is adopted.~%
Meanwhile, on the Totto dataset, \ourframework with the table reconstruction (i.e., \ourframework$_{w/\,\text{TRLoss}}$) does achieve the best performance in terms of BLEU-1, BLEU-2, METEOR, and ROUGE-L, though the improvement is not remarkable.~%
The contents of the input tables are mainly linguistic and the table structures are not too diverse might be able to explain the performance improvement of \ourframework$_{w/\,\text{TRLoss}}$ on the Totto dataset. 
With the above comparisons, we can conclude that, for the input tables with diverse structures, the limitation of the current text deliberation mechanism cannot be neglected if one aims to enhance the capability of table comprehension for the table-to-text task. 
Moreover, this also suggests that the generalization capability of text deliberation of \ourframework should be improved in the future.

\vspace{-1ex}
{\bf{Limitations.}}~%
In this work, long tables in the Totto dataset are removed since the efficiency and performance of \ourframework on large tables could be lowered. 
In the future, the capability of handling long tables for table-to-text models should be further explored. 
Besides, a large-scale and more exhaustive human evaluation is necessary. 
We plan to recruit more volunteers to conduct the human annotation.

\section{Conclusion} \label{sec:conclusion}

In this paper, to realize table-to-text with the pretrained language model, we proposed a table structure understanding and text deliberating approach, namely \ourframework. 
The table structure understanding was realized by developing a hierarchical multi-head attention network, which can benefit the fine-tuning of the text-to-text pretrained model.~%
The fully represented table information benefits not only the pretrained language model but also the text deliberation process since the structure information with rich semantics could be fed into the second-pass decoding naturally.~%
We carried out extensive experiments on two public datasets with different table types. 
Automatic and human-based evaluations, as well as qualitative analysis, validate the effectiveness of our approach to generating faithful and fluent table descriptions.
In the future, we will improve text deliberation by devising a unified framework to integrate the multi-pass decoder and refine the descriptive text paying more attention to sentence fluency.

\section*{Acknowledgements}
This work is supported in part by Foshan HKUST Projects (FSUST21-FYTRI01A, FSUST 21-FYTRI02A).

\bibliographystyle{acl_natbib}
\bibliography{myref}


\appendix

\begin{figure*}[b]
    \centering
    \includegraphics[width=\textwidth]{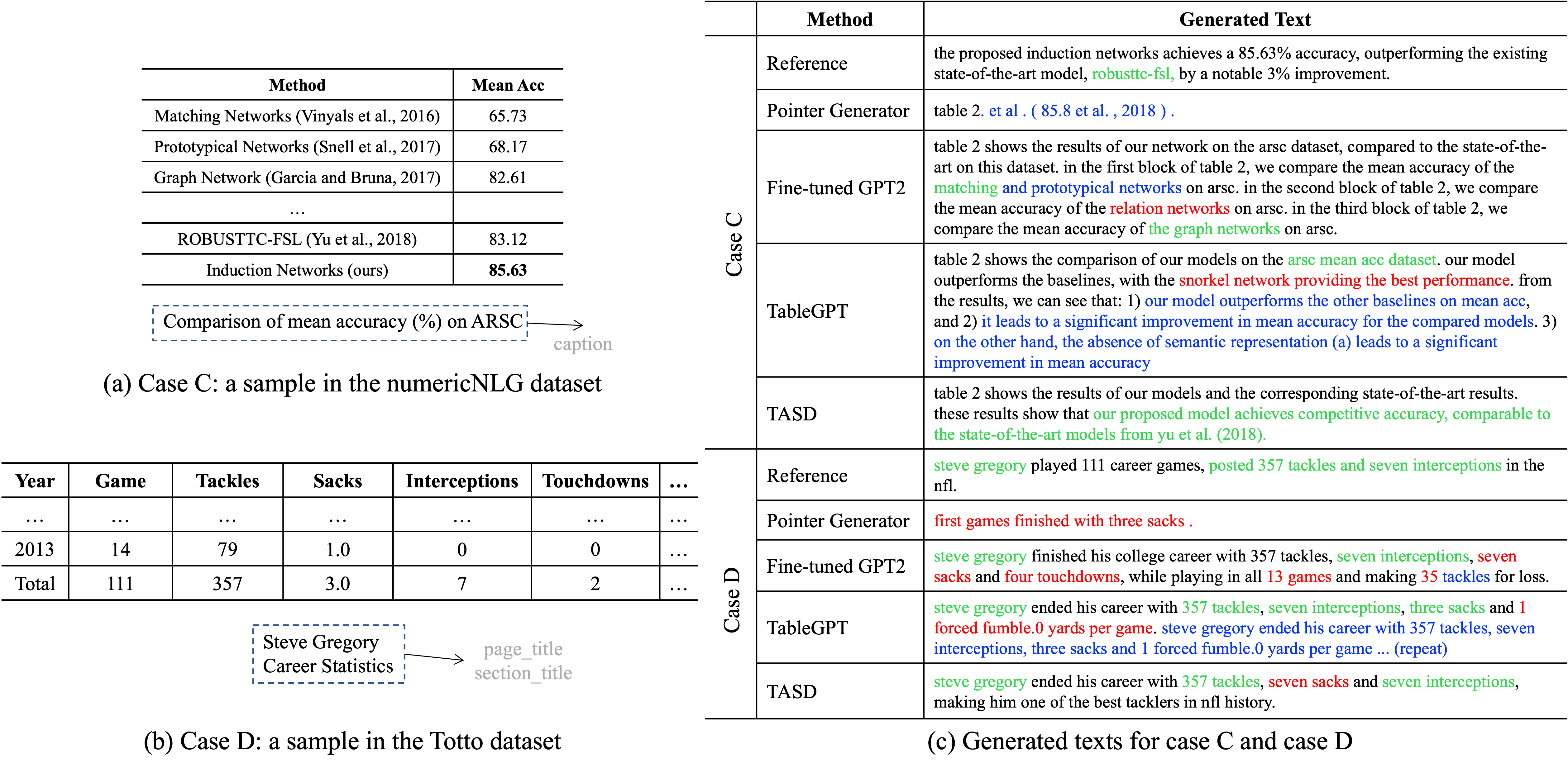}
    \vspace{-2ex}
    \caption{
    Generated table descriptions on cases C and D.
    }
    \label{fig:casecd}
\end{figure*}

\begin{figure*}[ht]
    \centering
    \includegraphics[width=\textwidth]{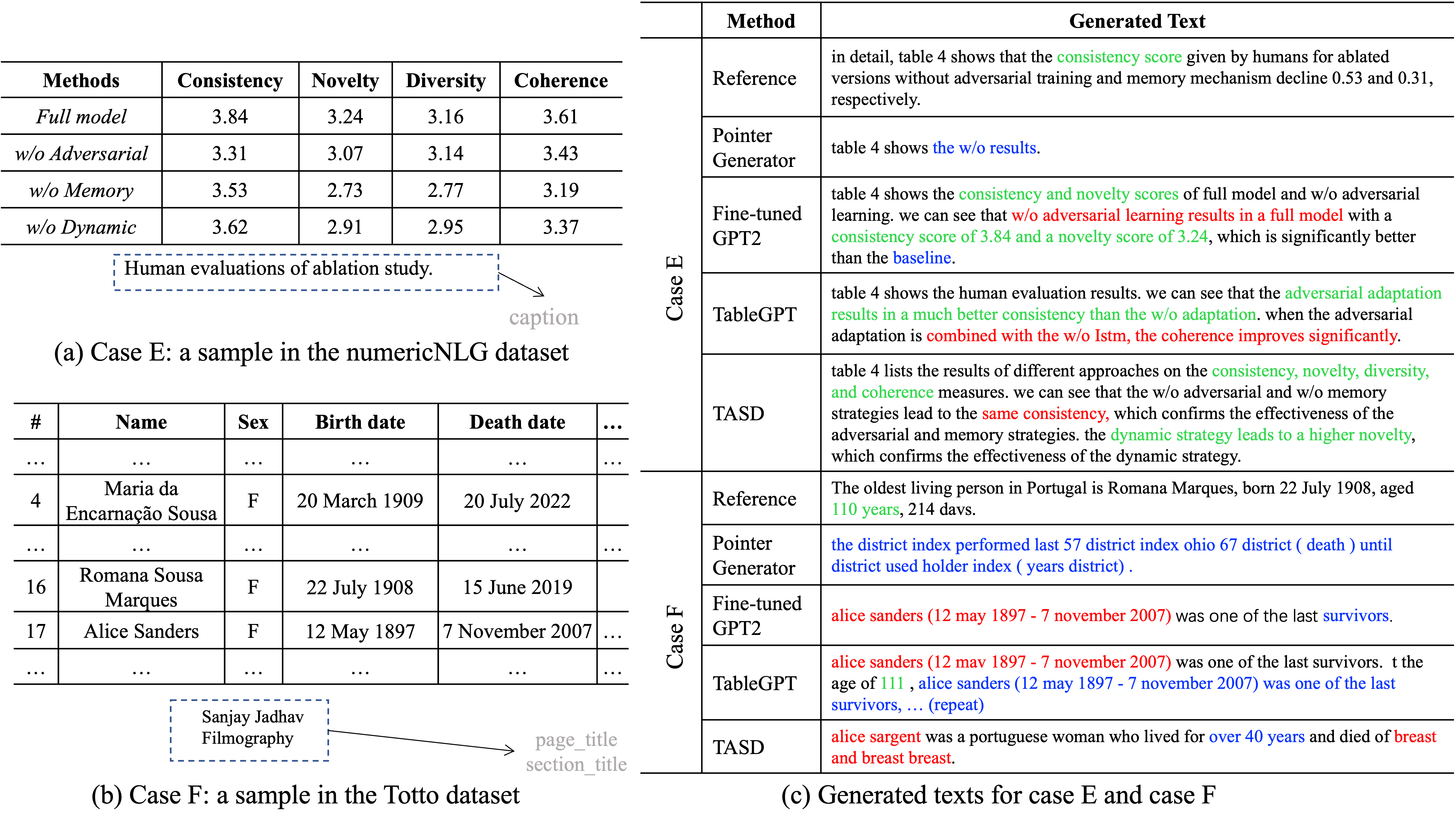}
    \caption{
    Generated table descriptions on cases E and F.
    }
    \label{fig:caseef}
    \vspace{30ex}
\end{figure*}


\section{Human Evaluation Settings}
\vspace{-1ex}

The criteria adopted in our human-based evaluation are 
(1) \textbf{Grammar} (e.g., is this paragraph grammatical?),
(2) \textbf{Coherence \& Concise} (e.g., is this paragraph coherent and contextually consistent? does it repeat redundant information?), 
and (3) \textbf{Factual perspective} (e.g., are the facts that this paragraph describes correct? are these facts related to references and tables?).
More specifically, we list the detailed justifications on how to score the generated text in each criterion as follows.  

\noindent
\textbf{Grammar}

\begin{enumerate}[leftmargin=0.5cm,label={$\bullet$}]
    \vspace{-1ex}
    \item 
    \textbf{1} 
    It is more like garbled code than a paragraph.
    \vspace{-1.5ex}
    \item 
    \textbf{2} 
    There are many obvious grammatical mistakes.
    \vspace{-1.5ex}
    \item 
    \textbf{3} 
    There are a few obvious grammatical mistakes.
    \vspace{-4ex}
    \item 
    \textbf{4} 
    There are few grammatical mistakes.
    \vspace{-1.5ex}
    \item 
    \textbf{5} 
    There are no grammatical mistakes.
\end{enumerate}

\vspace{-1ex}
\noindent
\textbf{Coherence \& Concise}

\begin{enumerate}[leftmargin=0.5cm,label={$\bullet$}]
    \vspace{-1ex}
    \item 
    \textbf{1} 
    The logic of text expression is chaotic and nonsense.
    \vspace{-1ex}
    \item 
    \textbf{2} 
    There are a lot of logical inconsistencies or redundant information.
    \vspace{-1ex}
    \item 
    \textbf{3} 
    There are some logical inconsistencies or redundant information.
    \vspace{-1ex}
    \item 
    \textbf{4} 
    There are a few logical inconsistencies or redundant information, but it does not affect browsing.
    \vspace{-1ex}
    \item 
    \textbf{5} 
    The logic of the text is smooth without redundant information.
\end{enumerate}

\noindent
\textbf{Factual Perspective}

\begin{enumerate}[leftmargin=0.5cm,label={$\bullet$}]
    \vspace{-1ex}
    \item 
    \textbf{1} 
    The paragraph does not coincide with the reference or table, and  it is full of information inconsistent with the facts.
    \vspace{-1ex}
    \item 
    \textbf{2} 
    The paragraph describes the facts incorrectly and has a low correlation with reference, but is related to the information in the table.
    \vspace{-1ex}
    \item 
    \textbf{3} 
    The paragraph description is incorrect, but it is highly coincident with the reference.
    \vspace{-1ex}
    \item 
    \textbf{4} 
    The paragraph description is basically correct, and the coincidence with the reference is low, but it also describes the information in the table.
    \vspace{-4ex}
    \item 
    \textbf{5} 
    The paragraph description is correct and highly coincident with the reference.
\end{enumerate}

\section{Illustrative Examples of Generated Descriptions}
\vspace{-1ex}

We additionally selected another two examples of the generated table descriptions from the numericNLG and Totto datasets, respectively.
The results are shown in \cref{fig:casecd,fig:caseef}. 
From these four examples, we can see that \ourframework can generate more accurate and fluent descriptive texts.
While incorrect descriptions can be found in the outcome texts generated by different models for cases D and F, which suggests that generating faithful descriptions for open-domain tables is much more challenging and requires more powerful and, thus larger, pretrained language models.

\section{Extra Implementation Details}
\vspace{-1ex}

The learning rate of GPT2 was searched from $\{3e-4, 3e-5, 3e-6\}$. 
In the evaluation of discussing the limitation of text deliberation (see \cref{sec:discuss}), a trade-off parameter for balancing the GPT2 fine-tuning loss and the TRLoss was adopted, then the trade-off parameter was searched from $\{1e-1, 5e-2, 1e-2, 5e-3, 1e- 3\}$, and 1e-2 was selected for the reported performance. 
Besides, the reported results in \cref{table:disscussion-eval:numericNLG,table:disscussion-eval:Totto} were averaged in 3 runs.


\end{document}